%% file: egpaper.tex
\documentclass[10pt,twocolumn,letterpaper]{article}

\usepackage{wacv}
\usepackage{times}
\usepackage{epsfig}
\usepackage{graphicx}
\usepackage{amsmath}
\usepackage{amssymb}
\usepackage{comment}
\usepackage[dvipsnames]{xcolor}
\definecolor{Mycolor2}{HTML}{00F9DE}
\wacvfinalcopy % *** Uncomment this line for the final submission

\ifwacvfinal
\def\assignedStartPage{1} % *** Enter the assigned starting page number (instead of 9876)
\fi

\ifwacvfinal
\else
\fi

\begin{document}

%%%%%%%%% TITLE
\title{Context-aware Padding for Semantic Segmentation}

\author{Yu-Hui Huang\\
KU Leuven\\
%{\tt\small yu-hui.huang@esat.kuleuven.be}
\and
Marc Proesmans\\
KU Leuven\\
%{\tt\small Marc.Proesmans@esat.kuleuven.be}
\and
Luc Van Gool \\
KU Leuven /\ ETHZ\\
%{\tt\small @esat.kuleuven.be}
}

\maketitle
%\thispagestyle{empty}
%
%Papers, excluding the references section, must be no longer than eight pages in length.
%%%%%%%%% ABSTRACT
\begin{abstract}
   %As the most common approach, z
  Zero padding is widely used in convolutional neural networks to prevent the size of feature maps diminishing too fast. 
  However, it has been claimed to disturb the statistics at the border~\cite{Nguyen19distribution}. As an alternative, we propose a context-aware (CA) padding approach to extend the image.
  We reformulate the padding problem as an image extrapolation problem and illustrate the effects on the semantic segmentation task.
  Using context-aware padding, the ResNet-based segmentation model achieves higher mean Intersection-Over-Union than 
  the traditional zero padding on the Cityscapes and the dataset of DeepGlobe satellite imaging challenge. 
  Furthermore, our padding does not bring noticeable overhead during training and testing. 
  %it does not bring too much overhead during training and testing. 
  %## del?
  %Compared to other methods, ours alleviate the gap to the true statistics on the border region.
\end{abstract}

%%%%%%%%% BODY TEXT

\section{Introduction}
%segmentation Introduction
Semantic scene segmentation is a fundamental task in computer vision. Its application ranges from autonomous driving to robot navigation. Since the success of the fully convolutional networks~\cite{Long15fully}, more and more techniques are developed to improve the segmentation accuracy, like pyramid pooling~\cite{zhao2017pyramid} or atrous spatial pooling~\cite{dilated_conv,chen2018deeplab}.

\begin{figure}[h]
\centering
\includegraphics[height=40mm]{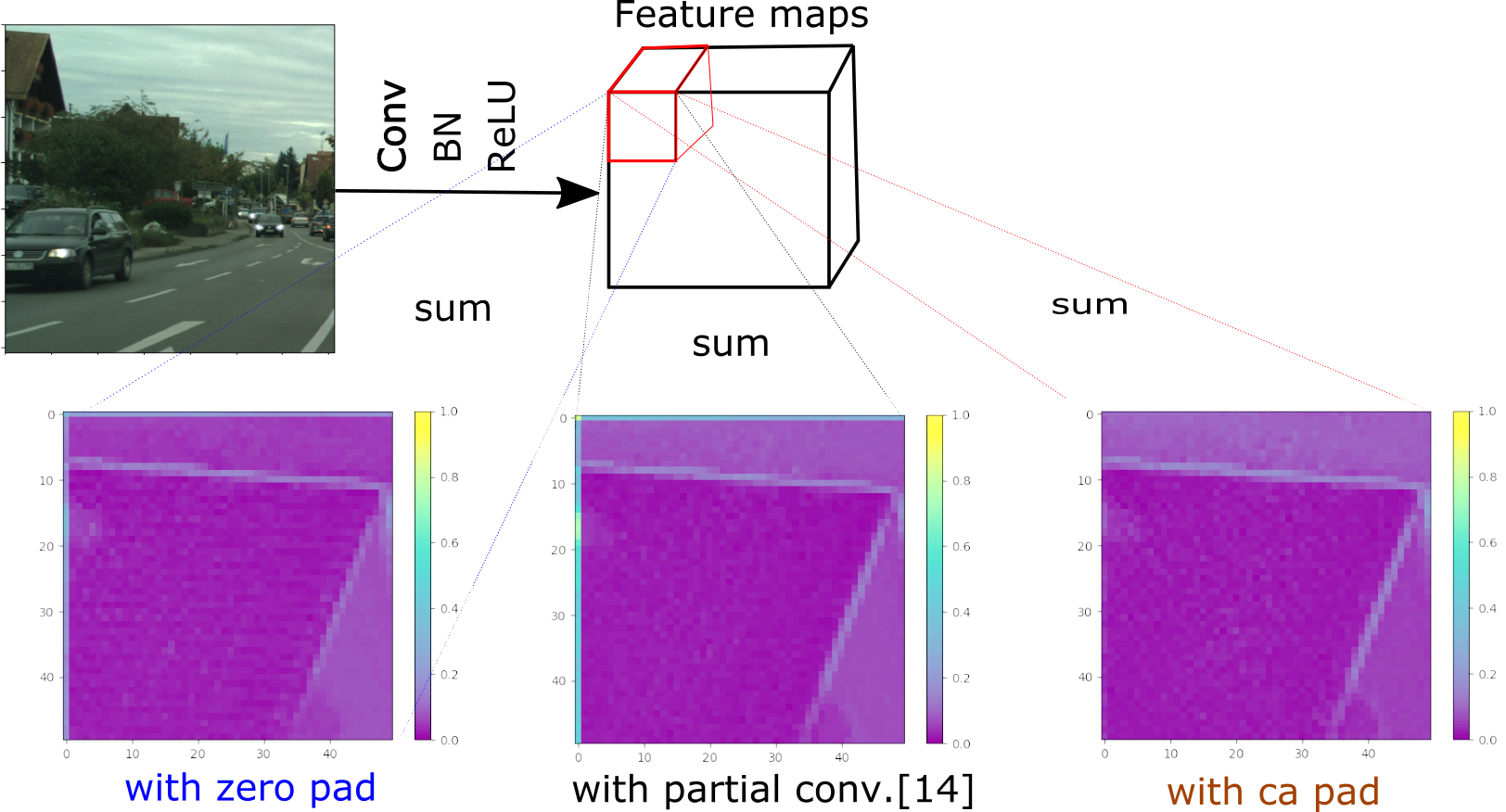}
\caption{
%Top left crop from an activation map (summed across channels) from the convolution with different padding methods. Compared to two other methods, CA-padding has more smooth boundary without any artifact on the left border of summed activation map. (\textbf{Best viewd by zooming in.}) %The activation map after ReLU is summed across the channels.
Visualization of a summed activation map from the convolution with different padding methods. We crop the top left corner of each one and normalize them to the same range. %There are some artifacts on the left boundary for zero pad and partial conv. 
Compared to two other methods, CA-padding has more continuous boundary without any artifacts on the top/left border of summed activation map. (\textbf{Best viewed by zooming in.})
}
\label{fig:activ_map}
\end{figure}

%The essential operations in a neural network are defined by the convolution layers, and typically padding is applied to prevent feature maps from diminishing.
Padding is applied in convolution layers to prevent feature maps from diminishing. It is added to the frame of an image to allow for more space for the kernel to cover the image.
In most of the cases, zero padding is 
%applied to keep the 
used for reasons of
efficiency. However, adding additional zeros will alter the distribution around the border region \cite{Nguyen19distribution} (Fig.~\ref{fig:activ_map}). Although 
%the ratio of padding compared to the whole image is relatively small, 
the padding area is relatively small compared to the whole image,
it may still deteriorate performance since the learnt filters are shared among all the spatial locations including the borders. The network needs to learn to adapt %adopt 
itself to cope with the plausible distributions at the border.
%segmentation
Different from classification tasks, the effect may be %more severe 
more prominent
for dense labeling tasks as we care about the prediction at every pixel.

%other paddings
%In order to eliminate the border statistic issue, 
In order to reduce border effects, 
one can apply other padding techniques such as reflection padding which reuses the value near the border in the reverse way or circular padding which has been applied in some specific setting~\cite{wang2018omnidirectional}. 
%In addition, there are some works focusing on the padding. 
Nguyen et al. \cite{Nguyen19distribution} proposed a distribution padding to maintain the statistics of the border region. 
%In a different manner, 
Alternatively, Liu et al. \cite{Liu18partial} proposed a re-weighting based scheme called partial convolution.
%based padding. 
Those methods contribute to a better performance in the classification and segmentation tasks, however, 
%% we argue that there is still some room for improvement.
we observe that one research path remains unexplored.
% -> unexplored
%

In this paper, we introduce a novel context-aware padding approach, in short CA-padding. We reformulate the padding problem as an image extrapolation task, and train a separate network to predict the area outside the image.
Using this approach, we intend to improve the existing methods by including more realistic statistics. Since the information in the image level will be propagated through layers to the end, we consider it is important to understand what the network receives in the beginning.
Therefore, we start with predicting the padding for the first layer of convolutional neural networks. 

%% moved above
%We reformulate the padding problem as an image extrapolation task. For that, we train a separate network to predict the area outside the image.

%outpainging
%In terms of image extrapolation task, generative adversarial networks (GAN) based models are popular choices.
In the context of image extrapolation, the use of generative adversarial networks (GANs) would naturally come to mind. They are quite popular as they have 
%been
shown to generate realistic results. There have also been works using multi-stage processes to improve the results even further.
Related approaches tackle the image extrapolation task %can be 
by means of autoregressive models. 
However, using such generative models in our context, would bring quite some computational overhead. These methods are typically based on a much more complicated architecture, and require longer inference time, thus are less attractive when using high resolution images.

Our method saves the computation time by only taking local region to directly predict the displacement to extrapolate the image. Furthermore, it does not require a heavy network and the computation is executed in the local region instead of the whole image and thus make it efficient. Also, for different spatial regions, we train separate networks to have them focused on the right information. At run time these models are executed in parallel as there is no dependency between them.

%CA-paddings
%Do not use any additional Latex macros.
We evaluate our proposed padding model on top of state-of-the-art segmentation models on two different semantic segmentation datasets. One for autonomous driving and one for satellite image recognition. 
The main contributions of this paper can be summarized as follows:
\begin{enumerate}
\item A simple neural network to perform image extrapolation as a padding method for the convolution layers.
\item The neural network based model behaves better than the naive padding approaches, since it exploits the context information and therefore its prediction is closer to the real statistics.
%predict the padding with the distribution closer to the real statistics.
\item We study the effects of the proposed padding method in the context of semantic segmentation, and compare with other methods.
%A deeper study of padding for semantic segmentation task.
\end{enumerate}

%===========================================================
\section{Related Works}
\label{sec:related}

In this section, we briefly introduce the related works in the following categories, semantic segmentation, padding and image extrapolation.
%\subsection{Boundary handling}
\subsection{Semantic segmentation}Starting from fully convolutional neural networks (FCN)~\cite{Long15fully}, the semantic segmentation task has been effectively tackled by deep neural networks. Following this work, researchers mostly focused on either integrating more context information in the network, which helps to disambiguate local information, or using local cues to arrive at better spatial predictions. The amount of research is vast and it includes the use of encoder-decoder architectures~\cite{badrinarayanan2015segnet,chen2018encoder}, dilated or atrous convolutions~\cite{dilated_conv,chen2018deeplab}, pyramid pooling modules~\cite{zhao2017pyramid}, squeeze-and-excitation modules~\cite{hu2017squeeze}, multi-scale predictions~\cite{eigen2015predicting,zhao2017icnet}, attention models~\cite{chen2016attention}, feature fusion~\cite{liu2015parsenet,pinheiro2016learning}, context modules~\cite{zhang2018context,yuan2018ocnet}, etc.
The reader can refer to~\cite{garcia2017review} for a review on some of  aforementioned semantic segmentation techniques.
At their core, however, all these approaches leverage the great representational power of deep models, like~\cite{resnet}, trained on large datasets of fully-annotated images, like~\cite{lin2014microsoft,cityscapes}, to achieve impressive results.
%\cite{pascalvoc2012}
In this paper, our focus is not the exact architecture for semantic segmentation. Rather we want to investigate the impact of padding on the segmentation performance.
Hence the underlying idea is as well that the proposed method can be applied to any of aforementioned models.

\subsection{Padding} In addition to zero padding, there are different kinds of padding applied such as circular padding or reflection padding under the context of signal processing. For reasons of efficiency, there is a tendency to use zero padding in convolutional neural networks 
%instead of
rather than other types of padding. In the recent years, padding gained some more interest though. %there are some various research about padding.
Liu et al.~\cite{Liu18partial} proposed the partial convolution and it was also applied to tackle the image inpainting task. It is done by re-weighting the results from convolution layers near the image border based on the ratio between the padding area and the convolution sliding window area. Later on, Nguyen et al.~\cite{Nguyen19distribution} proposed a mean distribution padding method to replace zero padding for classification and image generation task. In this way, they try to keep the same statistics on the border to avoid the border effect.
In the special case of even-sized kernels, Wu et al.~\cite{Wu19convolution} proposed to apply symmetric padding. % for even-sized kernels.
Triess et al. ~\cite{Triess20scan} utilized cyclic padding to provide more context when processing the LiDAR points.
As opposed to these methods, our method uses the context information to provide a better padding prediction.

\subsection{Image extrapolation}
Image extrapolation or image outpainting refers to the task of predicting the region beyond the borders of a given image.
Prior to the success of the deep neural networks, 
%The typical solution to image extrapolation task is done in data-driven manner. The 
the missing parts in the images were filled by retrieving similar patches from the input image~\cite{zhang2013framebreak,Kopf2012quality} or other images of a large database~\cite{Wang14bigger}.
%Zhang et al. \cite{zhang2013framebreak} extrapolate the field of view of a photograph by learning with a guide image.
Since the success of generative adversarial models (GANs), a few learning-based methods have been designed for indoor 3D structure~\cite{Im2pano3d} or human body images~\cite{portraitComp}.
Recently, Wang et al. \cite{Wang19wide} proposed a semantic regeneration network which first expands the feature and then predicts the extended image. 
Teterwak et al. \cite{Teterwak19boundless} improves the discriminator with semantic conditioning to generate a more semantic coherent extensions.
Yang et al. \cite{yang2019very} utilizes recurrent mechanism to keep the content consistent for long natural scenery image extension.
In addition to GANs based methods, there is another type of model for image generation, autoregressive models. Models such as PixelRNN~\cite{oord48pixel}, PixelCNN~\cite{oord17conditional} and so on are typical examples. 
These generative methods can generate plausible results, but, as indicated in the introduction, they would bring additional overhead (training effort, inference time, architecture size, etc) on their use in this particular padding setting.

In this paper, we provide a simple but accurate enough method for image extrapolation. It does not need multi-stage training and is easy to be integrated into e.g. a segmentation model.

%-----------------------------------------
%TODO params in fig.
\begin{figure*}
\centering
\includegraphics[height=60mm]{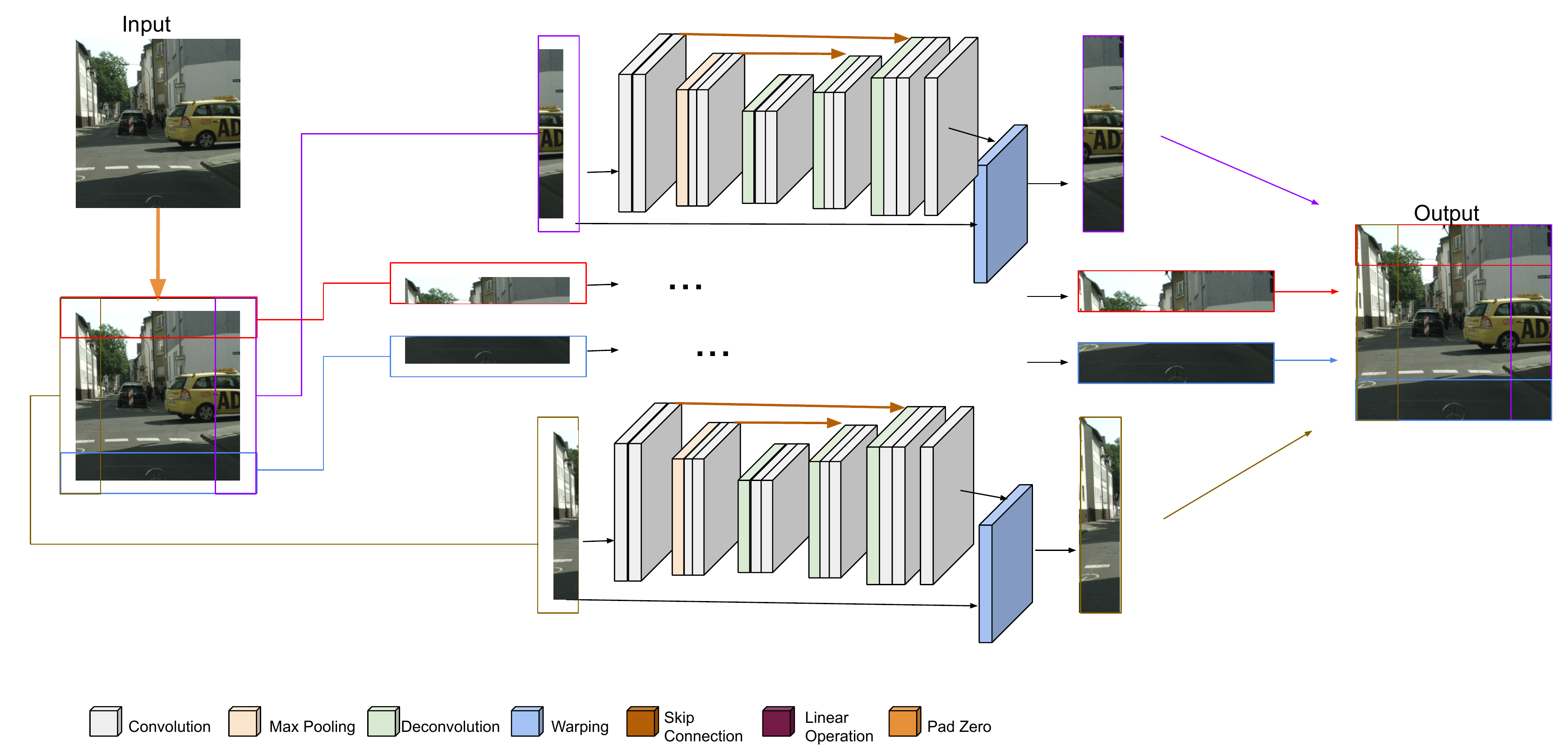}
\caption{
The pipeline of our context-aware padding model.
}
\label{fig:arch}
\end{figure*}
\section{Method}
\label{sec:method}
%## er
Here we explain our padding approach. As mentioned earlier, since padding is an essential component in a convolutional neural network, the additional overhead due to the padding process itself should be kept to a minimum, and hence we want to avoid that the practical use of the method would be compromised by the increase in training (and inference) time. Therefore we aim to find a balance between the accuracy of the padding model and its efficiency.

\subsection{Image level padding}
We model the image padding problem as a 'local' image extrapolation problem. Given an RGB image $I$ with the shape ($h,w$), the goal is to have a model $T$ to predict a spatialwise extended image $I'$ where the size of $I'$ equals to the shape of $I$ plus 
two padded areas on each of the vertical (height) or horizontal (width) sides of $I$.
%twice of padding numbers of the first convolution layer of the target network. 
Suppose 
%$p$ is 
we set the amount of padding in the convolution layer $l$ to $p$.
%where we aim to apply our padding for. It 
This is equivalent to taking an image of size ($h+2*p, w+2*p$) as the input to $l$ with no further padding during the convolution computation.

In order to save GPU memory,
%To make the memory efficient computation, 
we only use the regions nearby the border to predict the padding values. For each side of the image, we train separate padding prediction model. Take the left side for example. We first crop the left $m$ pixels from the image, append $p$ of zeros at the left side of the crop and thus form a new block $b$ with the size ($h, p+m$). The block $b$ serves as the input to our padding model.

%## check with figure 1 
Our padding prediction model is inspired by~\cite{liu2017voxelflow}, and is composed of two parts, 
\begin{itemize}
\item a displacement network, based on a simple encoder-decoder architecture with skip connections.
\item a warping layer. 
\end{itemize}
The architecture is also illustrated in Fig.~\ref{fig:arch}. %More details can be found in the supplementary document. 
Given a local region $b$ as input, the encoder-decoder $E$ 
%first 
predicts a pixelwise displacement $F$ which gives an idea on how the padding (border) area is correlated to the central part. Then the warping layer $T$ exploits the predicted displacements to fill in the padding area using the input pixels. 
%warping based on the predicted displacement.
\begin{equation}
F = E(b,W)
\end{equation}
where $W$ denotes the network parameters of $E$.
For every pixel $(x_i, y_i)$ in $b$, the coordinates w.r.t. the ones after transformation $(x_i^{warp}, y_i^{warp})$ are associated as,
\begin{equation}
x_i = x_i^{warp} - \Delta x_i, y_i = y_i^{warp} - \Delta y_i.
\end{equation}, where $(\Delta x, \Delta y)$ denotes the displacement vectors predicted from the decoder.

With the warping layer, the output image $b'$ is formulated as,
\begin{equation}
	{b'}_i = \sum_{k \in N(x_i, y_i)} b_k(1-|x_i - x_k|)(1-|y_i - y_k|),
\end{equation}, where ${b'}_i$ denotes the value of the $i$-th pixel at $(x_i^{warp}, y_i^{warp})$ in the output $b'$, and $N(x_i, y_i)$ is the 4-neighborhood region of the pixel at $(x_i, y_i)$ in the input $b$.

\subsection{Reconstruction Loss}
%During the training process, 
The reconstruction loss for the training process is defined as follows.
Suppose, after the data augmentation, we obtain an image crop $b$. This image crop serves as the ground truth for the image extrapolation. We then apply a binary mask $M$ to the image crop, to mimic the border effect. The resulting image is set to be the input of our padding model $T$. In the mask, we set the value of the padding region to 0 and 1 for other known region.
To train the model, we optimize the following reconstruction loss.
 \begin{equation}
  L = \frac{1}{N} \sum_{b \in D} (\| b \odot (1-M) - T(b \odot M, F) \odot (1-M)  \|)
  \label{equ:dt}
\end{equation}

where $D$ is the training set, $N$ is its cardinality.

\subsection{Feature level padding}
The method described for image level padding applies to feature level padding as well. %Differently, 
%Yet, whereas for image level padding the outer bounds are defined by default, for the features maps they are calculated during the normalization process.
%we need to perform some pre-processing steps to normalize the feature maps. 
To generate the ground truth feature maps for training, we extract intermediate features from a pre-trained segmentation network. We then apply the mask $M$ on top of it as described in the previous section to generate the input to the network.
However, %from an initial experiment, 
during the experiments
we found that adding additional CA-padding to feature level gives only marginal improvement which does 
%not worth 
not weigh up against the additional computation overhead. 
%Therefore, in this paper, we focus on the experiments of image level padding.

%\textit{Data augmentation}
\begin{comment}
\subsection{Network architecture details} Similar to other image extrapolation works, GANs in particular, we adopt an encoder-decoder architecture. 
%Different from them, 
%## what architectures? is there a reference or not?
On the other hand, our network is lighter and does not need multi-stage training. Therefore it allows to process larger images in real-time.
%thus it is possible to process larger images. 
The displacement network consist of three parts, an encoder, a bottleneck and a decoder. The encoder consists of one to two convolution units while the decoder contains the same number of deconvolution units. Each convolution unit contains a convolution, a batch normalization and a max-pooling layer. The deconvolution unit has a bilinear upsampling and a convolution layer. Between the corresponding layers in the encoder and the decoder, there is a skip connection. 
\end{comment}
%-----------------------------------------
\section{Experiments}
\label{sec:exp}

\textbf{Datasets} We use Cityscapes and the dataset from the DeepGlobe Land Cover Classification Challenge~\cite{DeepGlobe18} to test our method. 
\begin{itemize}
\item
In the Cityscapes dataset, there are 5,000 images with fine pixelwise annotation collected by driving a car in 50 cities in different seasons. Those images are divided into train/ validation/ test set with 2,975/ 500/ 1,525 number of images. %In addition, there are 20,000 coarsely annotated images, but in our experiment we only use the images with fine annotations. 
Each image 
%is
has a resolution of 1024 by 2048 pixels. %
%And there are 19 classes.
The dataset is annotated for 19 classes.

\item
The dataset from the DeepGlobe challenge consists of 1,146 high-resolution satellite images. There are 803 / 171 / 172 images in the training / validation / testing set. Each image has corresponding pixelwise class labeled image with seven different classes including urban land, agriculture land, rangeland, water, barren land and unknown class.
\end{itemize}
%
%## 
In the experiments, we adopted the datasets' training 
%set to train our model and evaluate it on the validation set.
set for training, and used the corresponding validation set for evaluation.
 The evaluation is done in single-scale without flipping the input image.
\\

\noindent \textbf{Metric} To evaluate the segmentation performance, we follow the convention to compute the mean of class-wise intersection over union (mIoU). %For ADE20K, we also use pixel-wise accuracy (Pixel Acc.)  To evaluate the quality of our image-level padding, we choose the peak signal-to-noise ratio (PSNR) and mean squared error (MSE).
%ADE 20k?
\\

\noindent \textbf{Training Details} We optimize the padding model using the Adam algorithm~\cite{kingma15adam} with learning rate 1$\times {10}^{-4}$ and (${\beta}_1,{\beta}_2) = (0.9,0.999)$. The batch size is 8 and we train it for  150 epochs. For data augmentation, we apply the same one as for semantic segmentation model training. To be more specific, we adopt random mirror, random resize between (0.5, 2), random rotation between (-10, 10) degrees, random Gaussian blur and random crop to a specific size.

For segmentation, we optimize the model with SGD with initial learning rate of 1e-2 and poly learning rate. To train the model, we apply data augmentation as described in the previous paragraph. For both datasets, we crop the image to 713 by 713.
To test the model, we follow the same strategy as PSPNet by using the overlapping crops of the size 713 by 713. %For the ablation study, we chose PSPNet with ResNet18 as the backbone to evaluate the model with different settings. 
For other padding models, we adopt the same setting except for replacing the padding. 

For comparison, we followed the steps described in~\cite{Liu18partial} to train the baseline model with partial convolution. 
%To be more specifically, 
That is, we replaced all the convolution layers with the partial convolution in the encoder. In the case of PSPNet, it includes the backbone architecture and the pyramid pooling module. %--add 0624
In addition, we implement the mean interpolation padding from~\cite{Nguyen19distribution} as another related work. For a fair comparison, we also include other padding methods (replication, reflection and bilinear interpolation) under the same setting as ours which is to replace the first zero padding of the network.
%--end add 0624--

\begin{figure}[h]
\centering
\includegraphics[height=35mm]{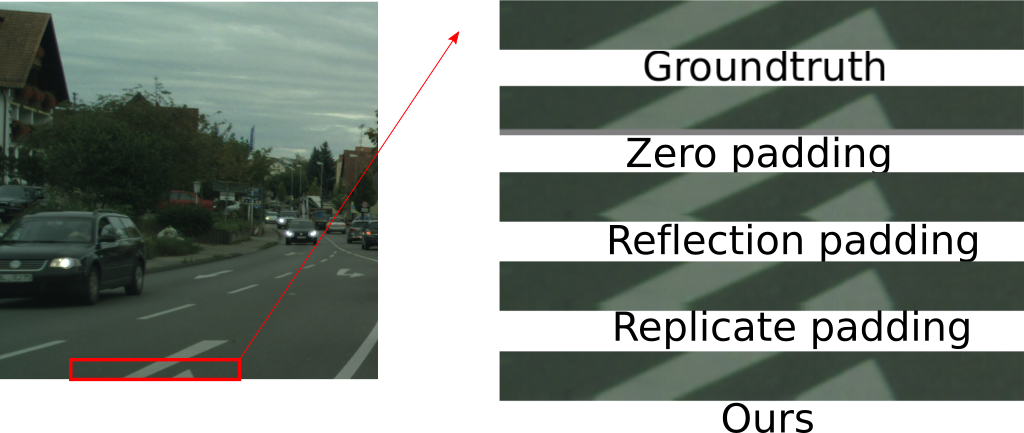}
\caption{
%Example of different padding methods. We zoom in for different padding cases of the red box from the image.
The results for the different padding methods is illustrated here on the cropped area within the red bounding box. The effect of the padding is visible in the bottom row of the images.
}
\label{fig:padf}
\end{figure}

\subsection{Padding evaluation}
\label{sec:padeval}
To simulate the padding in convolution layers during training the segmentation model, we use
%adopt 
the same data augmentation that is used for training the segmentation model itself. We apply overlapping crops and take the border region as ground truth to evaluate different padding methods.
We train our padding model using the training set and evaluate the padding performance on the validation set. We take an image and multiply it by a 0-1 mask to create the input of our model. The image before masking acts as the ground truth for our padding model.

To estimate the effect of our image padding model, we compare our models with other padding methods, including zero padding, circular padding, reflect padding, replicate padding and the padding by directly bilinear interpolation. Table~\ref{table:padevalade} and Table~\ref{table:padevalcs} presents the results on Cityscapes and the DeepGlobe dataset. For both dataset, we crop the input image to 719 by 719 pixels in a sliding window fashion. Each crop has one third overlapping with the previous crop. The notation $pad_i$ means we pad $i$ pixels at each side of image. In both cases, we take 20 pixels wide region to predict the padding. 
The results show that our method performs better than any of the other padding models
%From the result, we can see that our padding models outperform other padding models 
in both PSNR and MSE metrics. 
In general, the results of the $pad_3$ are always worse than $pad_1$ because the further it is to the known region the less information is provided. Among all other padding methods, %the 
replicate padding behaves best,
%acts more toward the ground truth 
but is still slightly worse than our model.

Fig.~\ref{fig:padf} illustrates an example of different padding. The left part shows an image after padding with 3 pixels at each side. In the right part, we zoom in the region specified by the red box and present the cases for different padding. We can see that from the reflection and replicate padding, the pattern of the sign becomes different. From this, we can infer that if there is object with specific structure on the border of the image, the reflection padding and replicate padding will change the pattern. %More examples can be found in the supplementary document.

%TODO Figure~\ref{fig:activ_map} presents 

\subsection{Segmentation Results}
We first compare our padding method with our state-of-the-art methods by replacing the padding method for the first convolution layer in the ResNet50. And the padding for other layers keep the zero padding. For comparison, we replace the convolution layers from the baseline with partial convolutions. Table~\ref{table:cs_psp50} presents the segmentation results in mIoU for different padding methods. From the table, we can see partial convolution improves the baseline zero padding method by 0.8\%. On top of that, our padding brings 0.1\% improvement more than partial convolution.  Though from the previous section the replicate padding shows good performance, it only brings 0.2\% improvement when it applies to the segmentation model. Among 19 classes, ours achieves the highest performance in 14 classes. To be noted that, for some classes of small object like \textit{pole},\textit{bike} and \textit{motorbike}, our padding method brings around 1 to 4\% improvement compared to the baseline. It might be because our padding method can generate the extension more closed to the real pattern rather than just filling in zero when an object lies on the border.

\begin{figure}[h]
\centering
\includegraphics[height=35mm]{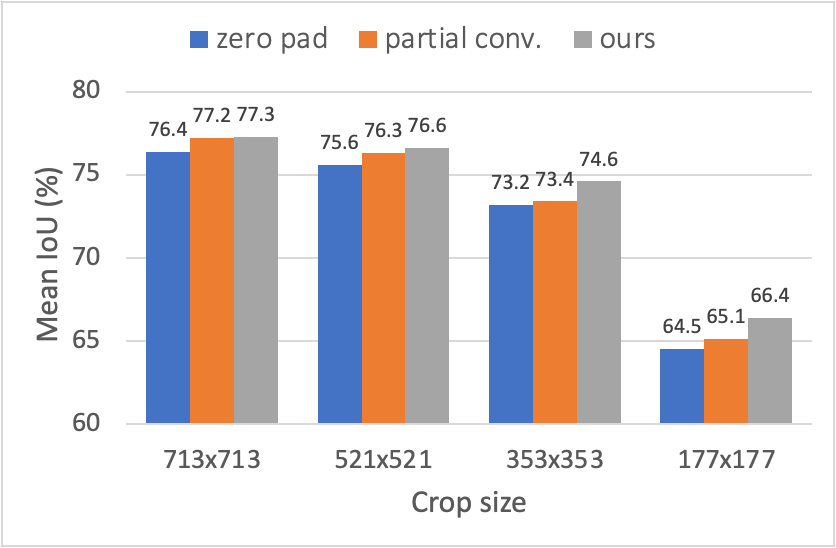}
\caption{
The segmentation result on Cityscapes val set with different padding models using different crop size during testing.
}
\label{fig:crop_plot}
\end{figure}

%new%
To evaluate how well our model improves at the boundaries, we test different crop size during evaluation. Fig.~\ref{fig:crop_plot} shows corresponding segmentation results. As we can observe from the figure, the smaller the crop size is, the more our CA-padding outperforms the baseline model (from +0.9\% to +1.8\%) due to more boundaries involved. %It shows our CA-padding benefits 

%%
%
% new table 1
%
\setlength{\tabcolsep}{4pt}
\begin{table*}[h]
\begin{center}
\scalebox{0.7}{
\begin{tabular}{llllllllllllllllllll|l}
\hline\noalign{\smallskip}
Method & road & swalk & build & wall & fence & pole & tlight & tsign & veg. & terrain & sky & person & rider & car & truck & bus & train & mbike & bike & mean\\
\noalign{\smallskip}
\hline
\noalign{\smallskip}
zero  & \textbf{97.9} & 83.6 & \textbf{92.1} & 51.9 &59.0 &62.8 &69.5 &77.0 &92.2 &63.7 &\textbf{94.4} &81.1 &60.6 &\textbf{94.7} &76.1 & 84.5 &71.4 &63.2 &76.6 & 76.4\\
%bilinear  & & &&&&&&&&&&&&&&& xx & xxx &xxx & xx\\
partial conv & \textbf{97.9} & \textbf{84.1} & \textbf{92.1} & 53.3 & 59.2 & 62.9 &69.6 & \textbf{77.3} & \textbf{92.4} &\textbf{66.6} &94.1 &81.3 &\textbf{60.9} &94.6 &\textbf{76.4} & 87.3 & \textbf{74.6} & 65.4 & 76.6 & 77.2\\
dist~\cite{Nguyen19distribution} & 96.9 & 77.9 & 90.0 & 41.4 & 52.9 & 56.5 &62.7 & 72.6 & 91.1 & 57.3 &93.3 & 77.3 & 52.5 &92.8 &62.3 & 71.5 & 62.0 & 60.4 & 74.1 & 70.8\\
replicate &97.7 & 82.1 & 92.0 & 55.4& 58.1 & 61.9 & 69.1 &76.5 &\textbf{92.4} & 65.4 &94.2 &81.2 &59.1 &94.5 &74.2 &86.8 &  72.9 & 64.1 & 76.4 &76.6\\
bilinear& \textbf{97.9} & 83.5 & \textbf{92.1} & 52.9 & 58.8 & 63.4 & \textbf{69.9} & 77.3 & \textbf{92.4} & 64.1 & \textbf{94.5} & 81.5 & 60.5 & 94.6 & 71.6 & 86.5 & 67.3 & 65.8 & 76.7 & 76.4\\
% xs model:
%ours & 97.9 & 83.5 &92.2 &54.6 &59.8 &63.4 &69.8 &77.3 &92.5 &65.9 &94.4 &81.3 &60.1 &94.9 &76.2 &87.6 & 77.3 & 62.0 & 77.1 & 77.2\\ 
%s model
ours & \textbf{97.9} & 83.5 &\textbf{92.1} &\textbf{56.2} &\textbf{59.5} &\textbf{63.5} &\textbf{69.9} &\textbf{77.3} &\textbf{92.4} &65.1 &\textbf{94.4} &\textbf{81.6} &\textbf{60.9} &94.6 &74.5 &\textbf{88.4} & 73.9 & \textbf{66.7} & \textbf{77.2} & \textbf{77.3}\\
%$\qquad\qquad$
\hline
\end{tabular}
}
\end{center}
 \caption{
Segmentation results in classwise IoU of the PSPNet with ResNet50 backbone on Cityscapes val dataset. "dist" denotes distribution padding and "bilinear" denotes bilinear interpolation.
}
\label{table:cs_psp50}
\end{table*}
\setlength{\tabcolsep}{1.4pt}

\begin{figure*}
\centering
\includegraphics[height=200mm]{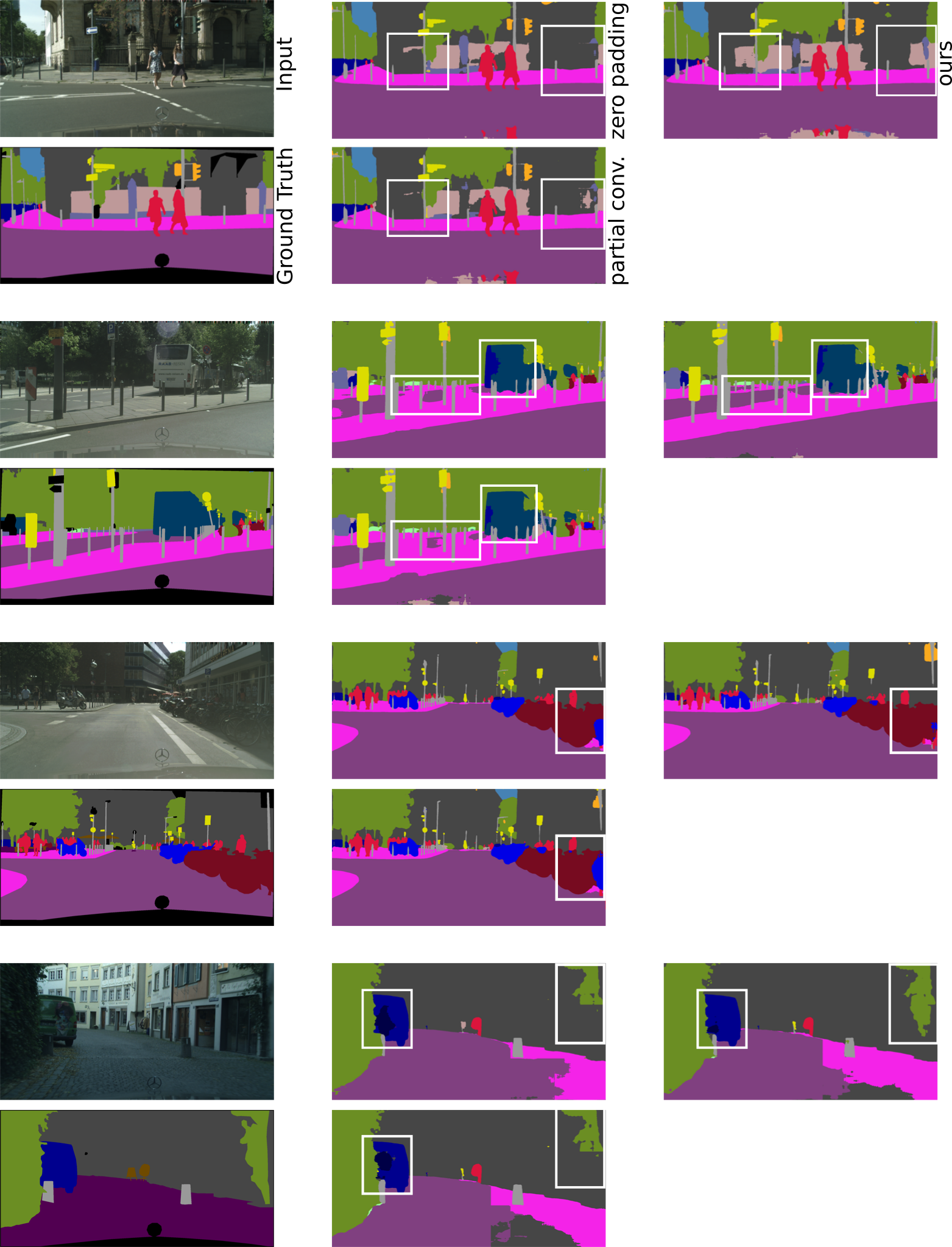}
\caption{
Qualitative results of PSPNet-ResNet101 with different padding, zero padding, partial convolution and our padding model.
}
\label{fig:example}
\end{figure*}

We further test our padding model with a deeper backbone, ResNet101, on Cityscapes. Table~\ref{table:cs_psp101} presents corresponding mean IoU for different padding methods. From the result, we can see our padding still improves the baseline by similar amount of mean IoU as ResNet50. It shows the eligibility of our model. Differently, the model with partial convolution has less improvement compared with the version with ResNet50 backbone. This shows that the weight rescaling method from partial convolution might disturb the learning when the network goes deeper. %
%
% new table 2 PSP 101
%
\setlength{\tabcolsep}{4pt}
\begin{table*}[h]
\begin{center}
\scalebox{0.7}{
\begin{tabular}{llllllllllllllllllll|l}
\hline\noalign{\smallskip}
Method & road & swalk & build & wall & fence & pole & tlight & tsign & veg. & terrain & sky & person & rider & car & truck & bus & train & mbike & bike & mean\\
\noalign{\smallskip}
\hline
\noalign{\smallskip}
zero  & \textbf{98.0} & 84.3 &92.4 &57.2  &59.3  &61.9  &70.2 &77.6 &\textbf{92.6} & \textbf{65.6} &94.4 & \textbf{82.2} &62.7 &94.8 &74.5 & 85.5 &76.3 &64.6  &77.5 & 77.4\\
partial conv. & \textbf{98.0} & \textbf{84.5} & 92.2& 54.5 &59.8 & 63.5 &70.7 &\textbf{78.0} &92.4 &63.9 &94.4 &82.0 &61.7 & \textbf{95.0} & \textbf{79.3} & 88.6 &\textbf{78.6} & 63.4 & 77.5 & 77.8\\
%& x &  & &  &  &  & & & & &  & &  &  &  &   &  &  &  & 77.53\\

% xs model:
%ours & 97.9 & 83.5 &92.2 &54.6 &59.8 &63.4 &69.8 &77.3 &92.5 &65.9 &94.4 &81.3 &60.1 &94.9 &76.2 &87.6 & 77.3 & 62.0 & 77.1 & 77.2\\ 
%s model
%ours &  & x &  & &  &  &  & & & & &  & &  &  &  &   &  & & \textbf{78.36}\\
ours & 97.9 & 84.3 & \textbf{92.5} & \textbf{59.7} & \textbf{60.4} & \textbf{64.1} & \textbf{71.0} &77.7 &92.5 &64.7 &94.4 &82.1 & \textbf{62.8} &94.9 &78.9 & \textbf{89.5} & \textbf{78.6} & \textbf{68.1} &\textbf{77.9} & \textbf{78.5}\\
%$\qquad\qquad$
\hline
\end{tabular}
}
\end{center}
\caption{
Segmentation results in classwise IoU of the PSPNet with ResNet101 backbone on Cityscapes val dataset.
}
\label{table:cs_psp101}
\end{table*}
\setlength{\tabcolsep}{1.4pt}
Fig.~\ref{fig:example} illustrates the segmentation results on Cityscapes validation set. The baseline zero padding model is trained with PSPNet-ResNet101. From the first row of the figure, the model with our CA-padding successfully predicts the fence on the left and right border of the image. Additionally, our model clearly delineates the road between the sidewalks from the example of the second row. The improvement is not only found on the borders but also closed to the center of image because the learnt filters are shared across the image. Therefore, a better padding method is helpful for learning better representation.

\begin{figure*}
\centering
\includegraphics[height=210mm]{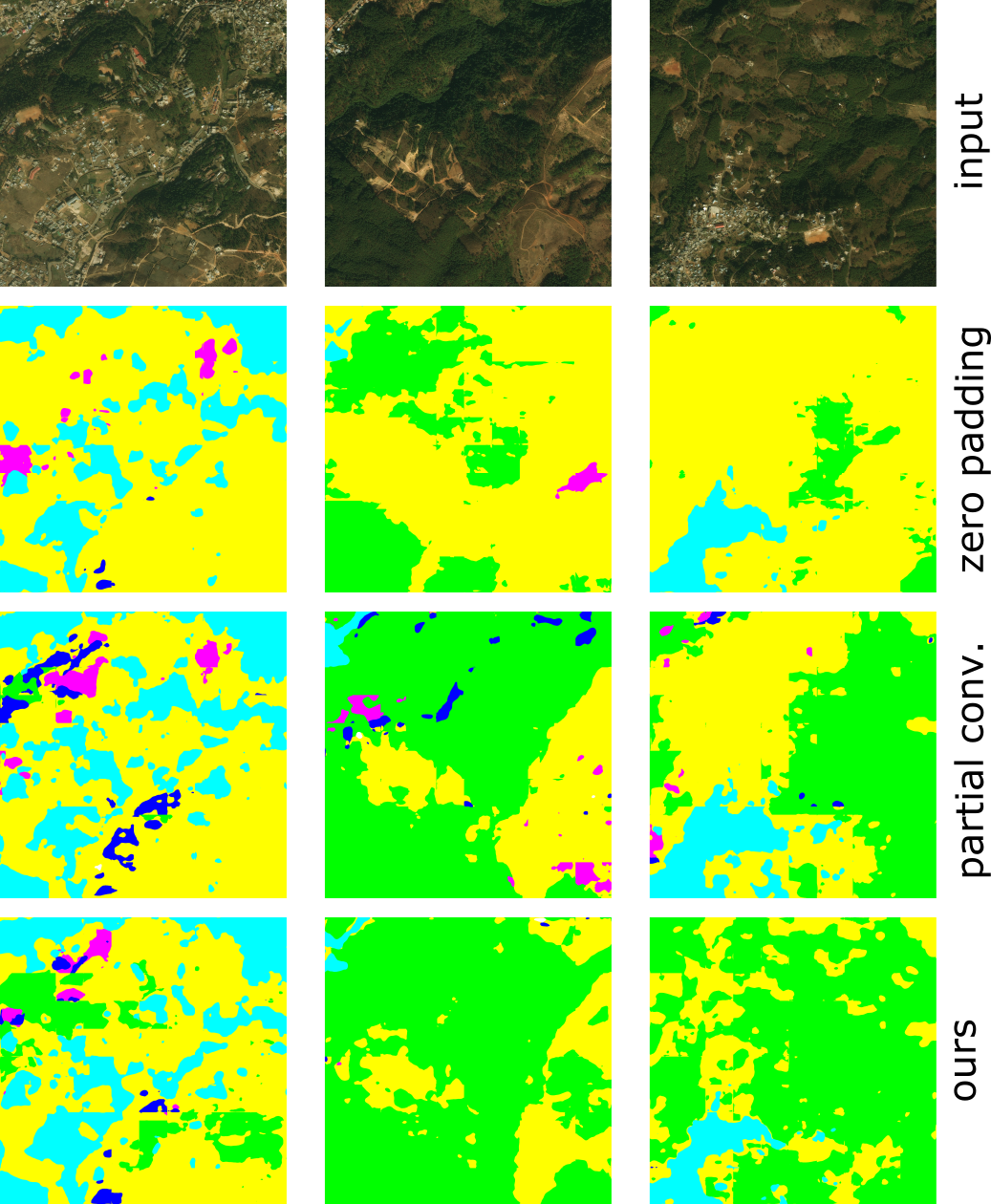}
\caption{
Qualitative results of PSPNet-ResNet18  on the DeepGlobe val set.  with different paddings : zero padding, partial convolution and our padding model. Colors: \textcolor{Mycolor2}{urban}, \textcolor{yellow}{agriculture}, \textcolor{purple}{rangeland}, \textcolor{green}{forest}, \textcolor{blue}{water}, \colorbox{lightgray}{\textcolor{white}{barren}}, unknown.
}
\label{fig:gexample}
\end{figure*}

To test our padding model in different type of scene other than driving, we perform experiments on a dataset from DeepGlobe Land Cover Classification Challenge. We take the PSPNet with ResNet18 as backbone as our baseline. Because the label of validation set is not available, we randomly choose 171 images from the training set to select our final model. Table~\ref{table:globeseg} presents the result on the official validation set received from the evaluation server. Our CA-padding outperforms the baseline zero padding method by around 2.5\% which is one to four percent more than partial convolution and replicate padding respectively. %new add
Fig.~\ref{fig:gexample} illustrates the segmentation results with different padding models on DeepGlobe validation set.
% new add
%
% new table PSP Resnet18 globeseg
%
%\setlength{\tabcolsep}{4pt}
\begin{table}
\begin{center}
\begin{tabular}{lll}
%\hline\noalign{\smallskip}
\hline
Method & mean IoU \\
%\noalign{\smallskip}
\hline
%\noalign{\smallskip}
zero padding  & 45.20  \\
partial conv. & 45.88   \\
distribution padding & 42.30 \\
replicate padding & 43.24 \\
bilinear interpolation & 42.03 \\
ours  & \textbf{47.84}   \\

\hline
\end{tabular}
\end{center}
\caption{
Evaluation of different padding method with PSPNet-ResNet18 on the DeepGlobe dataset validation set. 
}
\label{table:globeseg}
\end{table}
%\setlength{\tabcolsep}{1.4pt}
% TODO error reduce reason!
\begin{figure}
\centering
\includegraphics[height=40mm]{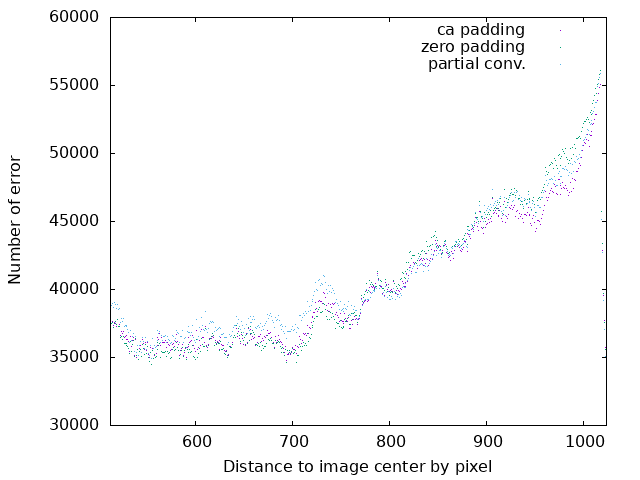}
\caption{
Error plot for the ResNet-101 on Cityscapes validation set. The x-axis is the distance to the image center and the y-axis is the number of error.
}
\label{fig:err}
\end{figure}

As indicated in section~\ref{sec:method}, it is possible to apply the CA-padding principle on feature level as well.
During the experiments
we found that adding additional CA-padding to feature level gives only marginal improvement. Whereas the CA-padding on image level increase the segmentation performance up to $77.3$, including additional CA-padding on the feature levels increases the performance up to $77.5$. As explained in section~\ref{ssec:time}, it would further add additional overhead on the computational time, proportional to the amount of layers.

%not worth 
%not weigh up against the additional computation overhead. 
%Therefore, in this paper, we focus on the experiments of image level padding.

\subsection{Error Analysis}
\label{sec:border}

To further understand the effect of our CA-padding on semantic segmentation compared to the zero padding, we plot the error distribution according to the distance to the image center. In this analysis, we take PSPNet with ResNet101 backbone as our baseline and evaluate the performance on Cityscapes validation set. We exclude the labels which do not count in the official evaluation (ignore labels). For each pixel of the prediction, we compare it with corresponding pixel from the ground truth label and create an error plot %. 
as shown in Fig.~\ref{fig:err} . 
Here we plot the distance of the pixels to the image center (x-axis) against the amount of pixels for which the ground truth labels and predicted labels differ.
Fig.~\ref{fig:err} illustrates this plot with three different settings, zero padding, CA-padding and partial convolution. It can be observed that the number of error achieves the maximum at around 1,000 pixel away from the center. 
%For the further part, the reason why the number of error drops might be due to many pixels belonging to ego vehicle or noises. 
We further zoom in the region from pixel 800 to 1,000, we can see that in most of the time the CA-padding (green dots) has least number of error. 
The second least is the partial convolution (blue dots) and then the zero padding (purple dots). From the result, we can infer that the distribution near the border region is improved by our CA-padding so that the number of error decreases compared with other padding methods.

\begin{figure}
\centering
\includegraphics[height=10mm]{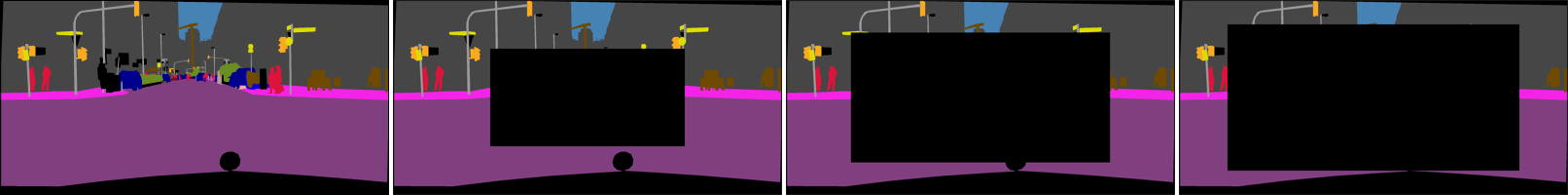}
\caption{
Examples of segmentation map with different leave-out proportion from the center regions.
}
\label{fig:frac}
\end{figure}

%To further 
Finally, in order to understand the performance close to the border region, we follow~\cite{Liu18partial} to evaluate the segmentation performance in the mean IoU metric on border regions. It is done by excluding the center regions when computing the mIoU. Fig.~\ref{fig:frac} illustrates examples of different proportions leaving out from the center regions. We set the center ($\frac{1}{3} \times \frac{1}{3}$, $\frac{1}{2} \times \frac{1}{2}$ ,$\frac{2}{3} \times \frac{2}{3}$, $\frac{3}{4} \times \frac{3}{4}$) to "don't care" labels in the ground truth. Table~\ref{table:frac} shows the results on the Cityscapes dataset. From the table, we can observe that when the leaving-out ratio goes higher, we obtain a better mean IoU. But %after 
when the ratio is larger than $\frac{1}{2} \times \frac{1}{2}$, the performance drops for all the cases. For the case of $\frac{3}{4} \times \frac{3}{4}$, both partial convolution and our padding improve the most compared to zero padding. The corresponding improvements are 0.43\% and 1.95\%.

\begin{comment}
\subsection{Generalizibility}
%We further investigate the difference in robustness of the model with different padding methods. 
In this section we investigate to what extend our model can be applicable for datasets for which it was not trained. 
%For that, 
As a use case, we took the model trained on Cityscapes training set and directly evaluated the performance on Grand Theft Auto V (GTA5) dataset~\cite{Richter_2016_ECCV}. The images from GTA5 were generated from a photo-realistic computer game named Grand Theft Auto V. There are in total 24,966 images with dense labelling which is compatible with Cityscapes. We choose the Part 1 which has 2,500 images to test the generalizability of the models. The classwise IoU can be found in Table~\ref{table:cs101_gta}. 
%From the result, we can see that the performance of all the models drop substantially. 
The table shows that the performance drops substantially for all models.
Yet, the model with our CA-padding method performs the best. It outperforms the zero padding and partial convolution by 3\% and 1\% respectively. Thus, we can conclude that the model trained with our CA-padding is more robust than the other padding methods.
\end{comment}
%

%
% new table 5 table:frac
% 
\setlength{\tabcolsep}{4pt}
\begin{table}[h]
\begin{center}
\begin{tabular}{l|lllll}
\hline\noalign{\smallskip}
Method & 0 & $\frac{1}{3} \times \frac{1}{3}$ & $\frac{1}{2} \times \frac{1}{2}$ & $\frac{2}{3} \times \frac{2}{3}$ & $\frac{3}{4} \times \frac{3}{4}$\\
\noalign{\smallskip}
\hline
\noalign{\smallskip}
zero          & 77.53 & 78.74 & 79.57 & 79.14 & 78.06\\
partial conv. & 77.88 & 78.95 & 79.63 & 79.27 & 78.49\\
ours  &         78.62 & 79.66 & 80.52 & 80.36 & 80.01\\

\hline
\end{tabular}
\end{center}
\caption{
Evaluation of different proportions of center region leaving out on Cityscapes dataset. Further details can be found in Sec.~\ref{sec:border}.
}
\label{table:frac}
\end{table}
\setlength{\tabcolsep}{1.4pt}

\subsection{Ablation Study}
%Here we explore the option of different design for our padding model. \\
Here we explore different designs for our padding model.
%\textbf{Network depth} To know the depth of the network has impact on the performance of our padding model, we train the padding model with a one-layer encoder and decoder. The performance 
%to train with the segmentation model. For the backbone with Res-50, the mean IoU drops around 0.1\% but for a deeper backbone like Res-101, the mean IoU drops 0.3\%. 
%In Table~\ref{table:paddepth}, we test our model with different depth of the network to see if it improves the accuracy.
\textbf{Size of input region} We further explore the size of input region to see if further context is useful to predict the displacement. Table~\ref{table:paddingregion} shows the MSE metrics with 20 and 30 pixels out of 713 pixels from the input. From the PSNR and MSE, given more region as the input seems not help. So we keep the design of 20 pixel to predict the padding.
\input{table_paddingregion}

\textbf{4-direction model needed?} In the design of our padding model, we separated the training in four different directions, left, right, top and down (cfr Fig.~\ref{fig:arch}). To test if it is necessary to train four models separately, we evaluate the padding only with the left and top model. In this setting, we use the left model to predict both left and right padding. The input from the right border is first flipped horizontally and sent to the left model. The same applies to the top model. We predict the top and bottom padding with the top model. We follow the same evaluation steps as in Section~\ref{sec:padeval} and calculate the PSNR and MSE. The corresponding numbers are 43.64 and 0.0002. Compared with the one trained with four different model from Table~\ref{table:padevalcs}, the PSNR drops around six and the MSE becomes ten times more. As a result, it proves the necessity of having four independently trained models for each direction.

\subsection{Training Time Overhead}
\label{ssec:time}
We measure the training time per epoch on a machine with four NVDIA V100 GPU for different padding. We take PSPNet with ResNet101 as our baseline (with all zero padding). On Cityscapes dataset, the baseline method takes about 233.873 second for one epoch training. After adding our CA-padding, the training time per epoch becomes 244.133 second. It increases 10.26 second for one epoch which is 4.3\% increase compared to the baseline. %--add 0924
Similar range of overhead applies to testing time as well. %--end add 0624
As a comparison, we also measure the training time for partial convolution. For one epoch, it takes 241.119 second which is a bit faster than our CA-padding but our method can bring further improvement. %Noted that if we further optimize our padding models by running them in parallel, the run time will decrease. 
Noted that the run time will decrease if we further parallelize our method.
\\

\textbf{Convergence Speed}
During training, we evaluate the model on validation set after each epoch. For each image in the validation set, we crop from the center a specific size. For total 200 epoch of training, we set a threshold on mean IoU to see which number of epoch can the model achieve. For the PSPNet-ResNet101 training on the Cityscapes dataset, the baseline model takes 148 epochs to achieve 76\% mIoU. The model with partial convolution and with CA-padding takes 136 / 117 epochs to achieve similar result separately. %From the result, we can 
It shows that our CA-padding can help converge faster than the baseline model and the model with partial convolution. 
To summarize, our model always achieves the highest performance, regardless the amount of epochs one chooses to train any of the alternative methods.

%
%
% new table 
%
\setlength{\tabcolsep}{4pt}
\begin{table}
\begin{center}
\begin{tabular}{lll|ll}
\hline\noalign{\smallskip}
      & ${pad_3}$ & & $pad_1$ & \\
model & $PSNR$ & $MSE$  &$PSNR$ & $MSE$\\
\noalign{\smallskip}
\hline
\noalign{\smallskip}
zero padding  & 29.10  & 3.9e-03 & 33.85 & 1.3e-03\\
bi. interpolation & 28.85 & 2.0e-04 & 29.92 & 5.5e-05\\
circular padding & 31.92 & 3.2e-03 & 36.66 & 1.1e-03\\
reflect padding  & 44.75 & 2.0e-04 &  53.27 & 2.1e-05\\
replicate padding  & 48.40 & 6.6e-05 & 57.82  & 7.5e-06\\
%$ours_{1-layer}$ & \textbf{49.27}  & \textbf{5.41e-05} & 58.64 & 6.4e-06\\
$ours$ & \textbf{49.17} & \textbf{5.51e-05} & \textbf{58.62} & \textbf{6.4e-06}\\
%$ours_{3-layer}$ & 49.15  & 5.53e-05 & \textbf{58.68} & 6.39e-06\\
%$\qquad\qquad$
\hline
\end{tabular}
\end{center}
\caption{Quantitative results for the padding prediction with padding size three and one on Cityscapes val set.
}
\label{table:padevalcs}
\end{table}
\setlength{\tabcolsep}{1.4pt}

\input{table_padevalade}
\section{Conclusions}
We proposed a context-aware (CA) padding method by transforming the nearby pixel to predict the padding value. Compared with the traditional generative models, our model is light-weight and easy to train. We explore the application of our padding model on semantic segmentation tasks. Compared with traditional padding methods, ours improves the state-of-the-art PSPNet on Cityscapes and a satellite imaging dataset. 

%For future work, we will investigate the possibility to re-design the overall approach, i.e. padding process and the segmentation task, in an end-to-end fashion. 

%\paragreph{\bf Acknowledgments}
\textbf{Acknowledgments}
We gratefully acknowledge the support of the TRACE project with Toyota TME.

%--------
\begin{comment}
\begin{figure}[t]
\begin{center}
\fbox{\rule{0pt}{2in} \rule{0.9\linewidth}{0pt}}
   %\includegraphics[width=0.8\linewidth]{egfigure.eps}
\end{center}
   \caption{Example of caption.  It is set in Roman so that mathematics
   (always set in Roman: $B \sin A = A \sin B$) may be included without an
   ugly clash.}
\label{fig:long}
\label{fig:onecol}
\end{figure}

\begin{figure*}
\begin{center}
\fbox{\rule{0pt}{2in} \rule{.9\linewidth}{0pt}}
\end{center}
   \caption{Example of a short caption, which should be centered.}
\label{fig:short}
\end{figure*}

%------------------------------------------------------------------------
\begin{table}
\begin{center}
\begin{tabular}{|l|c|}
\hline
Method & Frobnability \\
\hline\hline
Theirs & Frumpy \\
Yours & Frobbly \\
Ours & Makes one's heart Frob\\
\hline
\end{tabular}
\end{center}
\caption{Results.   Ours is better.}
\end{table}
\end{comment}
%-------------------------------------------------------------------------

{\small
\bibliographystyle{ieee_fullname}
\bibliography{egbib}
}

\end{document}

%% file: table_paddingregion.tex
%
% new table:paddingregion
%
\setlength{\tabcolsep}{4pt}
\begin{table}[h]
\begin{center}
\begin{tabular}{lll}
\hline\noalign{\smallskip}
size of input & PSNR & MSE \\
\noalign{\smallskip}
\hline
\noalign{\smallskip}
20 / 713 pixels & \textbf{49.27} & \textbf{5.41e-05}\\
30 / 713 pixels & 49.19 & 5.46e-05\\
%
%$\qquad\qquad$
\hline
\end{tabular}
\end{center}
\caption{
Quantitative results for the padding prediction with different range of region given as input to the network evaluated on the Cityscapes val set.
}
\label{table:paddingregion}
\end{table}
\setlength{\tabcolsep}{1.4pt}

%% file: table_padevalade.tex
%
%
% new table:padevalade
%
\setlength{\tabcolsep}{4pt}
\begin{table}[h]
\begin{center}
\begin{tabular}{lll}
\hline\noalign{\smallskip}
% & ${pad_3}$ & & ${pad_1}$ & \\
model & $PSNR$ & $MSE$ \\
\noalign{\smallskip}
\hline
\noalign{\smallskip}
zero padding  &  31.04  & 0.0036  \\
bilinear interpolation & 26.28 &0.0003  \\
circular padding & 35.16 & 0.0015   \\
reflect padding  & 41.79 & 0.0003  \\
replicate padding  & 44.00  & \textbf{0.0002}  \\
%$ours_{1-layer}$ & 40.19  & \textbf{1.0e-03} & \textbf{48.61} & \textbf{2.0e-04}\\
$ours$ & \textbf{44.48}  & \textbf{0.0002}  \\
%$ours_{3-layer}$ & \textbf{40.22} & \textbf{1.0e-03} & \textbf{48.61} & \textbf{2.0e-04} \\
%$\qquad\qquad$
\hline
\end{tabular}
\end{center}
\caption{
Quantitative results for the padding prediction with padding size three on the DeepGlobe val set.
}
\label{table:padevalade}
\end{table}
\setlength{\tabcolsep}{1.4pt}